\documentclass[dvipsnames]{article} %
\usepackage{colm2024_conference}

\usepackage{booktabs}
\usepackage{graphicx}
\usepackage{enumitem}
\usepackage{wrapfig}
\usepackage{algorithm}
\usepackage{algpseudocode}

\usepackage{microtype}
\usepackage{amsmath}
\usepackage{colortbl}
\usepackage[utf8]{inputenc}
\definecolor{lightgray}{rgb}{0.9,0.9,0.9}
\usepackage{caption}
\usepackage{subcaption}
\usepackage{setspace}
\usepackage{url}
\usepackage{multirow}
\usepackage{colortbl}
\usepackage{tabularx}
\usepackage{blindtext}
\usepackage{pgfplots}
\pgfplotsset{compat=1.18} 
\usepackage{tikz}
\usetikzlibrary{er,positioning,bayesnet}
\usepackage{makecell}
\usepackage{tipa}
\usepackage{siunitx}
\usepackage{nicefrac}
\usepackage{tocloft}
\usepackage{listings}
\usepackage[raster,skins]{tcolorbox} %
\usepackage{xltabular}
\usepackage{adjustbox}
\usepackage{xurl}
\usepackage{makecell}
\usepackage{booktabs}
\usepackage{multicol}
\usepackage{multirow}
\usepackage{rotating}
\usepackage{cleveref}
\crefname{section}{§}{§§}
\Crefname{section}{§}{§§}
\usepackage[normalem]{ulem}
\useunder{\uline}{\ul}{}

\usepackage{tablefootnote}
\usepackage{xcolor}

\usepackage{amsmath,amsfonts,bm}

\def\eqref#1{equation~\ref{#1}}

\def\1{\bm{1}}

\DeclareMathAlphabet{\mathsfit}{\encodingdefault}{\sfdefault}{m}{sl}
\SetMathAlphabet{\mathsfit}{bold}{\encodingdefault}{\sfdefault}{bx}{n}

\renewcommand{\hflink}{https://huggingface.co/Qwen/Qwen3-Coder-Next}
\renewcommand{\mslink}{https://www.modelscope.cn/models/Qwen/Qwen3-Coder-Next}
\renewcommand{\ghlink}{https://github.com/QwenLM/Qwen3-Coder}

\newcommand*\justify{%
  \fontdimen2\font=0.4em
  \fontdimen3\font=0.2em
  \fontdimen4\font=0.1em
  \fontdimen7\font=0.1em
  \hyphenchar\font=`\-
}

\newcommand{\model}{Qwen3-Coder-Next\xspace}

\renewcommand{\texttt}[1]{%
  \begingroup
  \ttfamily
  \begingroup\lccode`~=`/\lowercase{\endgroup\def~}{/\discretionary{}{}{}}%
  \begingroup\lccode`~=`[\lowercase{\endgroup\def~}{[\discretionary{}{}{}}%
  \begingroup\lccode`~=`.\lowercase{\endgroup\def~}{.\discretionary{}{}{}}%
  \catcode`/=\active\catcode`[=\active\catcode`.=\active
  \justify\scantokens{#1\noexpand}%
  \endgroup
}

\title{\model Technical Report}

\author{
\bf Qwen Team
}

\begin{document}

\maketitle

\begin{abstract}
We present \model, an open-weight language model specialized for coding agents. \model is an 80-billion-parameter model that activates only 3 billion parameters during inference, enabling strong coding capability with efficient inference. In this work, we explore how far strong training recipes can push the capability limits of models with small parameter footprints. To achieve this, we perform agentic training through large-scale synthesis of verifiable coding tasks paired with executable environments, allowing learning directly from environment feedback via mid-training and reinforcement learning. Across agent-centric benchmarks including SWE-Bench and Terminal-Bench, \model achieves competitive performance relative to its active parameter count. We release both base and instruction-tuned open-weight versions to support research and real-world coding agent development.
\end{abstract}


\begin{figure}[ht]
    \centering
    \includegraphics[width=0.8\linewidth]{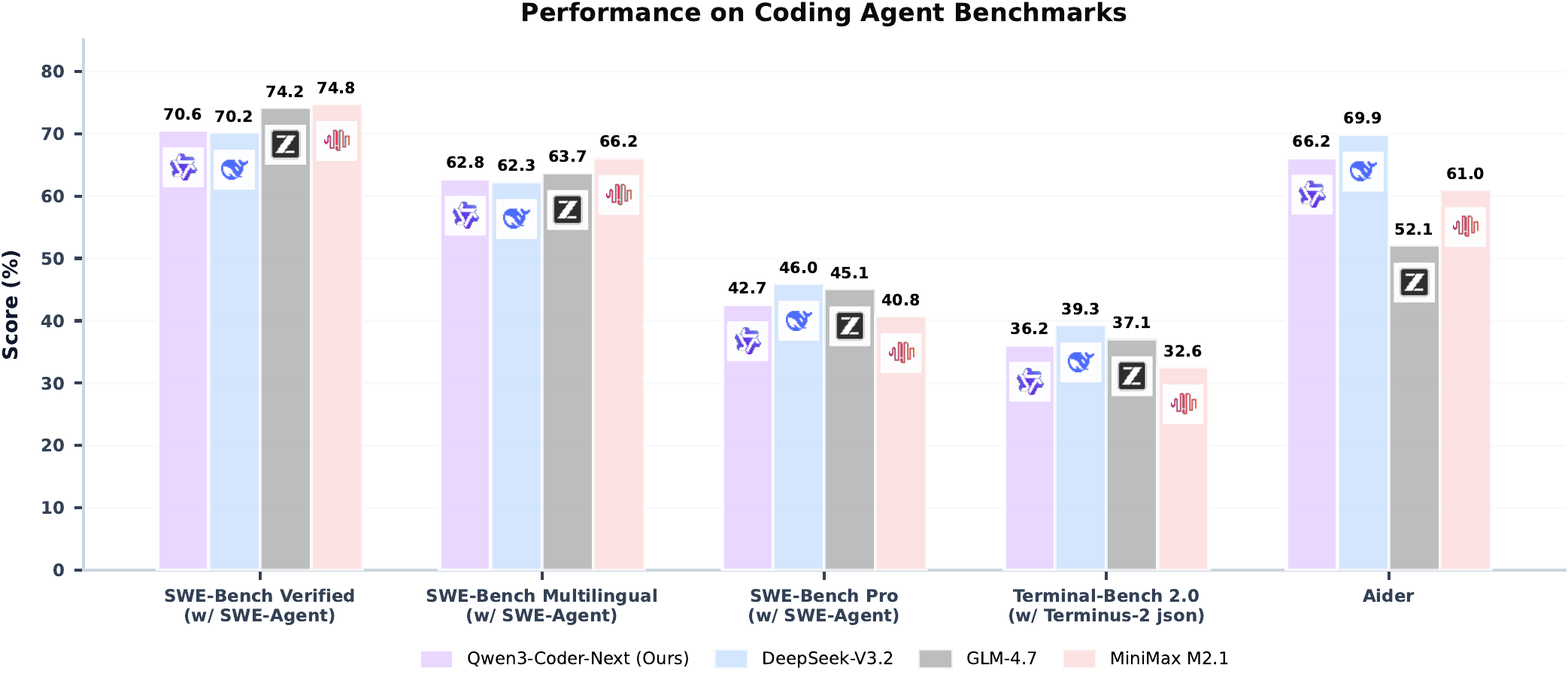}
    \caption{Comparison of \model to other open-weight models on SWE-Bench Verified, SWE-Bench Multilingual, SWE-Bench Pro, Terminal-Bench 2.0, and Aider.}
    \label{fig:benchmark_scores}
\end{figure}

\section{Introduction}
\label{sec:intro}

We introduce \model, an open-weight language model based on Qwen3-Next with hybrid attention and Mixture-of-Experts (MoE) designed specifically for coding agents and local development. 
It contains 80 billion total parameters while activating only 3 billion per forward pass, enabling fast inference and low deployment cost. Despite its lightweight active footprint, \model delivers strong performance across a wide range of coding benchmarks and real-world developer workflows.

The central advancement behind \model is the ability to scale agentic training. Modern coding agents must reason over long horizons, interact with real execution environments, and recover from cascading failures across multiple steps. Training for this regime requires more than static code data, which requires large volumes of verifiable, executable, and interaction-rich training signals~\citep{copet2025cwm,zhang2025agent}. To address this, we build a large-scale agentic training stack that synthesizes executable tasks, constructs reproducible environments, and learns directly from execution feedback. This stack enables reliable large-scale rollout collection and learning from real environment outcomes, supporting both mid-training adaptation and reinforcement learning for agent behaviors such as multi-step code editing, tool usage, and fault recovery in realistic development settings.


We train \model using a staged pipeline that progressively builds agentic capabilities without sacrificing base model generality. Starting from the pretrained base of Qwen3-Next, we first shift representations toward code- and agent-centric domains via continued pretraining. Next, we apply supervised fine-tuning on high-quality agentic coding data. From this checkpoint, we specialize multiple expert models, covering software engineering workflows, QA, web development, and user experience (UX) focused coding, and then distill their capabilities back into a single unified model. The result is an efficient, deployable model that delivers expert-level agentic performance across domains.

Extensive evaluation shows that \model achieves strong performance relative to its active parameter footprint. As illustrated in Figure~\ref{fig:benchmark_scores}, \model reaches competitive SWE-Bench Pro \citep{deng2025swebenchproaiagents} performance, outperforming or matching several models with an order of magnitude larger active compute. Beyond SWE-Bench Pro, we evaluate \model on an extensive suite of benchmarks spanning coding agents, general coding tasks, and general knowledge and reasoning tasks, with detailed results reported in Section~\ref{sec:experiments}. This efficiency makes \model particularly well-suited for production coding agents, where latency, throughput, and cost are first-order constraints. More broadly, these results suggest that scaling agentic training, rather than model size alone, is a key driver for advancing real-world coding agent capability.


\section{Scaling up Agentic Training}

Scaling agentic training to large volumes, across both mid-training and reinforcement learning, requires addressing two core challenges. First, we need a reliable pipeline for synthesizing verifiable tasks paired with fully executable environments. Second, we need an execution infrastructure capable of running a massive number of such tasks with high throughput and returning environment feedback efficiently. In this section, we describe our approach to each of these challenges, with a focus on task synthesis at scale.

\subsection{Task Synthesis}
\label{sec:task_synthesis}

We develop two complementary approaches for generating verifiable tasks with executable environments. The first approach grounds tasks in real-world software engineering issues by mining GitHub pull requests (PRs) and constructing corresponding runnable environments. The second approach starts from existing open-source datasets that already provide executable environments and synthesizes new task instances within them. Together, these methods enable large-scale task diversity with consistent execution-based validation.

\paragraph{Creating Executable Environments from GitHub PRs.}

To synthesize realistic software engineering tasks, we mine issue-related PRs and construct executable environments that reflect real-world bug-fixing tasks. After removing instances that overlap with downstream benchmarks, we decompose each PR into a buggy state, a corresponding fix, and an associated test patch. A specialized environment-building agent then constructs a runnable Docker environment and verification script, which is required to reliably distinguish the buggy and fixed states through execution. This process produces a large collection of verifiable software engineering tasks grounded in real repositories. 

Environment construction poses significant challenges for agents, leading to failure modes where agents exploit superficial verification shortcuts.
To mitigate this, we apply automated detection to identify and filter non-functional verifiers, and train a dedicated model to improve environment construction quality. We apply this model at scale to recent GitHub data to generate a large corpus of verifiable software engineering tasks, with all environments stored as reusable Docker images. To ensure dataset quality, we further employ a quality-assurance agent to automatically identify and remove ambiguous tasks, inconsistent environments, and misaligned tests before final inclusion. Further technical details can be found in \citet{chen2026sweuniverse}.

\begin{figure}[htbp]
\centering
\includegraphics[width=0.9\textwidth]{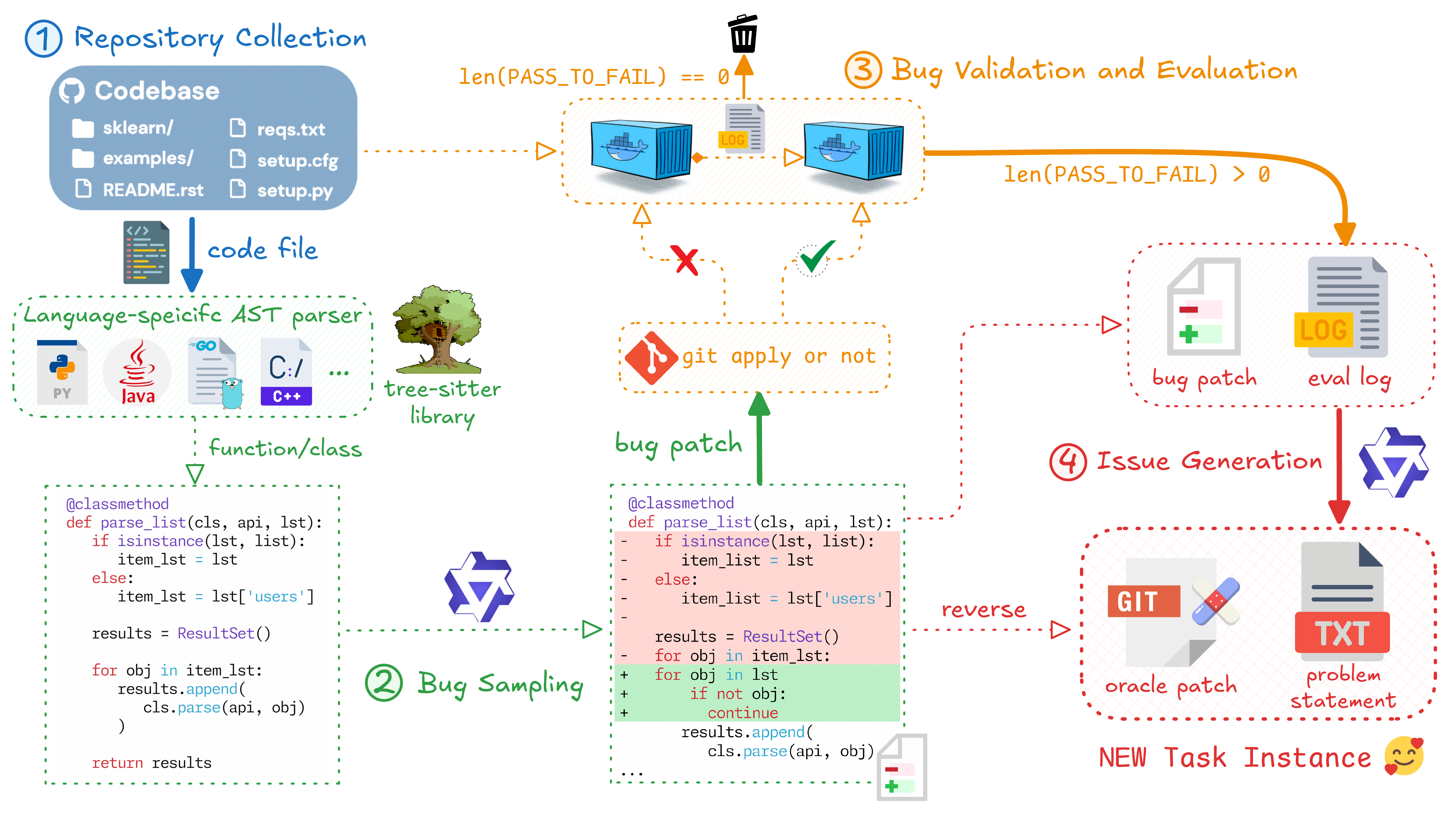}
\caption{Our pipeline for synthesizing bugs to scale up the number of software engineering tasks.} 
\label{fig:agent_env_2_and_3}
\end{figure}

\paragraph{Synthesizing Issues.}

In parallel, we synthesize additional software engineering tasks by building on prior work, including SWE-Smith~\citep{swe-smith}, SWE-Flow~\citep{swe-flow}, SWE-Rebench~\citep{swe-rebench}, and Multi-SWE-RL~\citep{multi-swe-bench}. These projects provide a strong foundation of seed tasks with executable repositories, test suites, and evaluation scripts. We extend these datasets to generate a substantially larger and more diverse set of verifiable software engineering problems. Our pipeline is shown in \Cref{fig:agent_env_2_and_3}.

Our synthesis pipeline systematically introduces controlled bugs into existing codebases and produces matching issue descriptions.
Building on curated, containerized repositories, we inject bugs via model-driven rewriting, semantic perturbations, and rule-based transformations, generalizing prior techniques to multilingual codebases. 
We retain generated bugs only when they fail existing tests and are resolved by patch reversion, guaranteeing that each task is both meaningful and tractable. 
To mitigate shortcut learning, we generate natural-language issue descriptions and exclude bug-triggering test files. 
This process yields approximately 800K verifiable software engineering task instances spanning over nine programming languages.

\subsection{Infrastructure}
\label{sec:infra}

Scaling agentic coding requires large-scale parallel execution and fully reproducible execution environments. To this end, we develop MegaFlow~\citep{megaflow}, an internal orchestration system. Our system adopts a fully cloud-native execution framework based on \textit{Alibaba Cloud Kubernetes}\footnote{\url{https://www.alibabacloud.com/en/product/kubernetes}}, enabling production-scale training, evaluation, and data generation for agentic coding workloads. In MegaFlow, each agentic coding task is expressed as an \textit{Argo workflow} composed of three logical stages: \textit{agent rollout}, \textit{evaluation}, and \textit{post-processing}. During the rollout stage, a single pod typically co-locates the agent container with the execution environment container (and additional auxiliary services when needed), enabling efficient long-horizon interaction with minimal communication overhead. The evaluation stage performs automated verification in a dedicated container, while the post-processing stage handles result interpretation, including parsing outcomes, extracting metrics, and optionally downstream analysis.

\section{Mid-training}
We now describe the mid-training stage used to specialize \model for coding and agentic tasks. Starting from the pretrained Qwen3-Next base model~\citep{qwen3next}, we perform targeted mid-training to adapt the model toward code reasoning, repository-level understanding, and agent-style interaction patterns.

\subsection{Data}
The guiding principle for data selection in mid-training is to balance natural and synthetic data. Natural data improves the model's general intelligence and robustness, but it does not fully match the distribution of tasks and interaction patterns observed in real user workflows. In contrast, heavy reliance on synthetic data can significantly improve performance on targeted tasks, but may lead to over-specialization, reduced response diversity, and weaker adaptation to other tasks during fine-tuning.

Therefore, our goal is to introduce the minimum amount of synthetic data required for the model to reliably perform common user tasks, while preserving response diversity and maintaining strong general-purpose capabilities. Following this principle, the mid-training corpus is composed primarily of natural data, supplemented with a smaller but carefully designed portion of synthetic data. We describe the key components below.

\subsubsection{Natural Data}\label{sec:natural_data}
Our sources of natural data include large-scale source code from GitHub and text–code grounding data derived from Common Crawl and targeted web domains. The pretraining corpus is updated through Sep 30, 2025.

\paragraph{GitHub.} Compared to the Qwen2.5-Coder series~\citep{qwen2.5coder}, we significantly expand coverage by increasing programming language support from 92 to 370 languages. We also incorporate substantially more pull requests, repositories, and code review data, and restructure these data sources to better reflect real development workflows.

We first include file-level pretraining data, which enables the model to learn strong file-level structural understanding. However, real-world software development rarely operates at the single-file level. Therefore, we place additional emphasis on repository-level code, allowing the model to learn cross-file dependencies and broader contextual relationships. To support this shift, we expand the training context length from 32,768 tokens to 262,144 tokens. Consistent with Qwen2.5-Coder, we use special tokens to concatenate repository data. In addition, we experiment with multiple repository serialization formats to improve generalization across different project layouts. In this release, repository-level data is expanded to approximately 600B tokens, representing a major portion of the mid-training recipe and proving more impactful than file-level datasets alone.

\paragraph{Text–code Grounding Data.} Text–code grounding data is collected from Common Crawl and domain-targeted sources such as math, programming, and education. We note that natural web data varies significantly in quality. Low-quality web content may contain incorrect information, insufficient context, or excessive code-switching between languages and formats. To mitigate these issues, we prompt Qwen3-Coder-480B-A35B-Instruct to rewrite web documents into normalized, structured text. The rewriting process removes advertisements, irrelevant HTML elements, and formatting artifacts, producing clean Markdown-style documents suitable for training. We empirically evaluate the impact of reformatting during mid-training. As shown in Table~\ref{tab:Reformat}, reformatting substantially improves our model across multiple evaluation benchmarks.

\begin{table}[ht]
    \centering
    \begin{tabular}{cccc}\toprule
         Model&  Evalplus & MultiplE&CRUX-Eval\\ \midrule
         Baseline&  54.38& 36.02&57.13\\
         Reformat&  63.09& 48.35&58.94\\
         \bottomrule
    \end{tabular}
    \caption{Impact of web document reformatting during mid-training.}
    \label{tab:Reformat}
\end{table}








\paragraph{GitHub Pull Requests (PRs).}

We construct PR-based training data by mining real-world GitHub pull requests and converting them into structured software engineering tasks. Each instance consists of a natural-language problem description, repository-level code context, and corresponding code edits. Problem descriptions are sourced from linked issues when available, or from PR titles and descriptions otherwise. Code context is reconstructed by reverting the PR patch and retrieving additional relevant files from the repository, intentionally introducing realistic mixtures of signal and noise. Edits are represented using both Search-and-Replace and standard \texttt{git diff} formats to support diverse editing paradigms. We then apply filtering and benchmark decontamination, removing anomalous files and instances overlapping with downstream tasks. This formulation encourages the model to localize bugs and produce precise code edits grounded in natural language descriptions.


\subsubsection{Synthetic Data}
Synthetic data is used to better align the model with real-world user workflows. We divide these tasks into two categories: 1) single-turn query, and 2) multi-turn agentic coding, where users interact with executable environments to complete tasks.

\paragraph{Single-turn QA.}
To improve single-turn QA capability, we use Common Crawl documents as seed data and prompt Qwen3-Coder-480B-A35B-Instruct to generate multiple grounded question–answer pairs per document. Generated questions must be self-contained and progressively increase in semantic depth or reasoning complexity. When documents lack sufficient quality or coherence, the model is allowed to abstain from generating QA pairs, reducing hallucination risk.

We also experimented with rewriting documents into alternative formats such as Wikipedia-style pages. However, this sometimes introduced hallucinated references or URLs. To avoid reinforcing these behaviors, we restrict rewriting to transformations that preserve original document content.

\paragraph{Multi-turn Agentic Coding.}

For multi-turn agentic data, we leverage synthetic tasks described in Section~\ref{sec:task_synthesis}. Trajectories are generated using multiple agent frameworks, including SWE-agent~\citep{sweagent}, Mini-SWE-agent~\citep{minisweagent}, OpenHands~\citep{openhands}, Claude-Code~\citep{claudecode}, Qwen-Code~\citep{qwencode}, and Terminus~\citep{merrill2026terminalbenchbenchmarkingagentshard}. We use Qwen3-Coder-480B-A35B-Instruct as the teacher model.

After generation, trajectories undergo strict rule-based filtering, including removal of missing termination signals, task failures, malformed tool calls. This produces a large-scale dataset of high-quality multi-turn tool-calling trajectories.

\begin{figure}[htbp]
    \centering
    \includegraphics[width=0.98\linewidth]{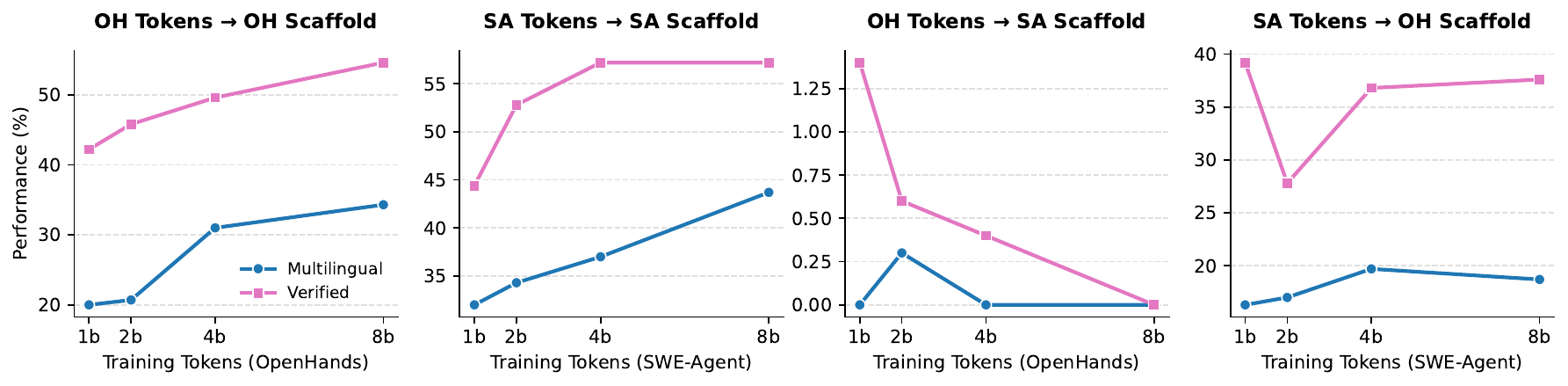}
    \caption{Scaling analysis of code agent pretraining on SWE tasks, SWE-Bench Verified and SWE-Bench Multilingual. \textbf{Left:} Within-scaffold scaling shows consistent improvement with the increase of mid-training tokens. \textbf{Right:} Cross-scaffold transfer reveals limited generalization across different agent frameworks.}
    \label{fig:scaling_law_for_cpt}
\end{figure}

To study scaling behavior, we analyze the relationship between mid-training token volume and downstream performance across agent scaffolds. As shown in Figure~\ref{fig:scaling_law_for_cpt}, we observe that within the same scaffold, performance consistently improves with increased mid-training tokens, demonstrating the effectiveness of large-scale agentic pretraining. Second, cross-scaffold transfer remains limited. Models trained on trajectories from one scaffold do not transfer strongly to others. Third, framework specialization plays an important role. For example, OpenHands, which is highly specialized for SWE tasks, transfers poorly to SWE-Agent, while transfer in the opposite direction is moderately successful. This highlights a trade-off between framework generality and specialization.

\subsubsection{Instruction-Following Data}
Because mid-training is dominated by natural documents, instruction-following behavior may not emerge reliably during mid-training alone. Therefore, we mix a small amount of instruction-following data into mid-training to enable early monitoring of downstream task performance during mid-training.

\subsubsection{Fill-In-the-Middle Code Completion}
\model supports fill-in-the-middle (FIM) code completion, which improves tasks such as document editing and online code modification \citep{chen-etal-2024-jumpcoder}. Using Stack-V2 \citep{starcoder2}, we synthesize FIM data in two formats: 1) chat-FIM, which embeds FIM tokens inside ChatML format \citep{ccc}, and 2) search-and-replace FIM, which generates diff-style patches.

To improve usability, we provide a proxy server that converts search-and-replace outputs into standard autocomplete-compatible formats. Experiments show that search-and-replace FIM outperforms Chat-FIM at an equivalent scale, likely due to strong alignment with PR-style pretraining data.\footnote{Detailed analysis is presented in \citet{SRI}.}

\subsection{Training}
In this stage, we train the model on trillions of tokens drawn from the mixture described above. To support multi-turn agentic trajectories, we extend the context length beyond typical pretraining settings to 262{,}144 tokens. In addition to standard next-token prediction, we also train the model using fill-in-the-middle (FIM) objectives, which are important for code editing tasks within long contexts.

We adopt best-fit packing (BFP)~\citep{best-fit-packing} as our sample packing strategy to avoid introducing context hallucination and head-side truncation when constructing combined document samples. Our BFP implementation achieves nearly the same efficiency as the traditional concatenate-then-split strategy during document index construction. For extremely long documents that exceed the model context length, we pre-split them into chunks matching the maximum input length. A pilot study comparing different packing strategies (see Appendix~\ref{app:best_fit_packing}) shows that, by trading off a negligible number of trailing padding tokens, BFP provides stable performance gains, particularly on long-horizon tasks.

Another challenge arises from the persistent presence of redundancy and noise within pretraining contexts, which can significantly reduce training efficiency. For example, while code headers and configuration blocks provide important contextual signals, repeatedly training on identical or near-identical patterns is redundant. To mitigate this, we apply masking to highly repetitive segments, so that we avoid potential repetitive behaviors of language models.

\section{Post-training}

\subsection{Supervised Fine-tuning}
After mid-training, we first perform supervised fine-tuning (SFT), which serves as the alignment stage bridging base model capabilities and complex human instructions. While much of the underlying knowledge is acquired during pre-training and mid-training, SFT reshapes this knowledge into instruction-following behaviors and improves response consistency across diverse interaction settings.

\paragraph{Data Composition and Sources} We curate a high-quality SFT dataset from three primary sources to ensure broad coverage and strong technical depth: in-house proprietary corpora, consisting of high-quality data accumulated from internal research and development, with a focus on alignment quality and safety behaviors; verified agentic trajectories, consisting of step-by-step action sequences validated through execution; and documentation-grounded open-domain QA, consisting of large-scale open-ended questions with emphasis on coding-related tasks, where candidate answers are filtered based on functional correctness and security.





\paragraph{Filtering with Verification.} We deploy a specialized agent model using Mini-SWE-agent to perform verification. This agent acts as a user simulator. Given a response from the assistant, the simulator attempts to execute the proposed code or commands from an end-user perspective. It evaluates system feedback signals, such as compiler outputs, runtime errors, and environment state changes, to determine whether the response meaningfully advances the task or resolves the user's request. This closed-loop verification process allows us to filter hallucinated or non-functional solutions, substantially increasing the density of executable and reasoning-valid training data.

\paragraph{Preference Modeling via Pairwise Judging} In addition to functional verification, we apply pairwise preference evaluation to refine conversational quality and response style. For each user request, we sample $n$ candidate responses using our strongest in-house models, forming $\binom{n}{2}$ unique candidate pairs. These pairs are evaluated by a dedicated pairwise judging model trained to score responses against a multi-dimensional checklist, including factual accuracy, task usefulness, and conversational style. The judge performs detailed comparisons to produce an ordinal ranking across candidates. Fine-tuning on data ranked through this process leads to consistent improvements in: stylistic consistency across diverse task types, linguistic clarity and professionalism in open-ended interactions, and proactive engagement, including anticipating follow-up user needs and driving task completion.

\subsection{Expert Models}

To further specialize the model toward important real-world domains, we train a set of expert models targeting specific capability clusters. While all experts share the same initial model, they differ in data sources, training recipes, and evaluation.

\subsubsection{Web Development Expert}
The Web Development expert targets full-stack web coding tasks, including UI construction, component composition, and interactive behavior implementation. High-quality training data in this domain must satisfy both visual correctness, functional correctness requirements, and optionally artistic tastes.

\paragraph{Data.} To curate high-quality WebDev alignment data, we employ a multi-stage filtering pipeline. All code samples are rendered in a Playwright-controlled Chromium environment. For framework-based samples such as React, we first deploy a Vite server to ensure all dependencies and components are correctly initialized before evaluation.

We perform two complementary evaluation stages. First, static visual evaluation uses a Vision-Language Model (VLM) to judge rendered pages using high-resolution screenshots. The VLM evaluates layout integrity, content completeness, and UI quality using a structured checklist \citep{plotcraft}. Samples that fail visual quality checks or contain rendering artifacts are discarded.

Additionally, dynamic interaction evaluation verifies functional behavior through browser automation. We parse DOM trees to identify interactive elements and use an in-house model to generate task-oriented user actions such as clicking, form entry, or menu navigation. These actions are executed automatically, and the VLM compares pre- and post-action screenshots to verify correct page behavior. Samples exhibiting broken interactions or unstable state transitions are removed.

\paragraph{Training.} The WebDev expert is trained using filtered execution-valid WebDev trajectories, emphasizing consistency between visual appearance and runtime behavior.

\subsubsection{User Experience Expert}
We observe that standard software-engineering tasks (e.g., fixing GitHub issues) do not fully capture the challenges of real-world agentic coding in CLI/IDE settings. 
We thus complement these evaluations with several in-house benchmarks and optimize the model based on their feedback. In particular, we find that different CLI/IDE scaffolds (e.g., Cline, Qoder, OpenCode, etc) adopt distinct tool-calling schemas, which poses a substantial challenge for models to reliably follow tool-call formats. Motivated by this, we propose targeted optimizations for tool-call format adherence.

\paragraph{Data and Tool Calling.}
Our data come from diverse sources, spanning over many scaffolds, task types, programming languages, development frameworks, and user interaction patterns, and include both synthetic and real-world trajectories. Building on this collection, we conducted extensive data-cleaning and data-mixture ablations under the guidance of in-house benchmarks. 

The inherent diversity of the data sources introduces substantial noise, making data cleaning particularly challenging. Beyond generic filtering (e.g. no finish action, resource failure, tool call failed), we found that rule-based validation on tool-call format correctness is particularly effective for improving agentic coding performance in CLI/IDE settings. Enforcing tool-call correctness can raise the performance upper bound by preventing models from learning malformed instruction-following patterns. In addition, this improves agent efficiency by reducing invalid tool calls and retries. These observations motivate us to treat tool-call format as an important objective during training.

\begin{figure}[htbp]
    \centering
    \includegraphics[width=1.0\linewidth]{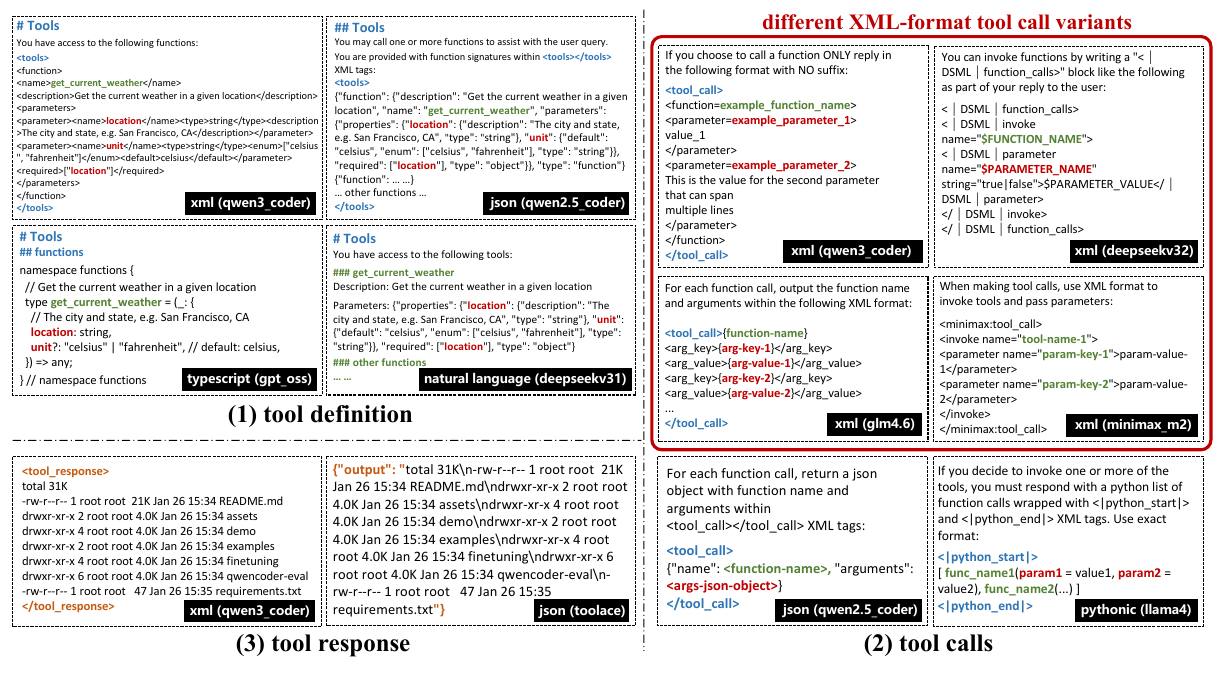}
    \vspace{-1em}
    \caption{Different tool chat templates, split by tool definition, tool calls, and tool response.}
    \label{fig:tool_chat_template_scaling}
\end{figure}

Many existing models are trained with a single tool chat template, which often leads to overfitting to specific output structures and reduced robustness when deployed under unseen tool-calling formats. In practice, real-world agent systems use a wide variety of formatting conventions, and users frequently define custom tool-call schemas directly in system prompts. To improve generalization, we train the model using diverse tool chat templates and formats.

As illustrated in Figure~\ref{fig:tool_chat_template_scaling}, tool chat templates differ along several key axes, including tool set definition, tool invocation format, and tool response wrapping. While JSON is a widely used protocol, it often introduces heavy escaping overhead for multi-line code. To address this, we also introduce an XML-style tool calling format, \texttt{qwen3\_coder}, which is designed for string-heavy arguments and allows the model to emit long code snippets without nested quoting.

Our training data incorporates a wide range of tool representations, including natural language tool descriptions, JSON-based formats, Python-style calls, XML-style schemas (including \texttt{qwen3\_coder}), and TypeScript-style interfaces. By exposing the model to diverse formatting conventions during training, the model learns format-invariant tool-use behavior rather than memorizing a single output structure.

\begin{figure}
  \centering
  \includegraphics[width=0.4\textwidth]{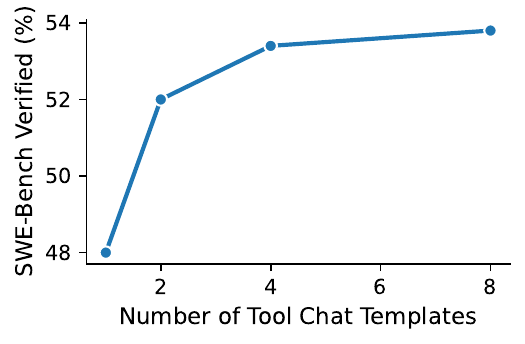}
  \vspace{-1em}
  \caption{SWE-bench Verified performance vs. number of tool chat templates. Data volume and training configuration are kept identical.}
  \label{fig:format_scaling}
\end{figure}

Empirically, increasing the number of tool call templates used during training consistently improves downstream robustness to format variation. As shown in Figure~\ref{fig:format_scaling}, performance on SWE-bench Verified improves as template diversity increases, even when the data volume and training recipe remain fixed. These results indicate that format diversity during training is an effective way to improve generalization to new tool-calling formats at deployment time.

\paragraph{Evaluating tool call format following.} Different IDE/CLI frameworks, such as Qwen-Code, Trae, OpenCode, Cline, and KiloCode, adopt customized prompt templates together with distinct function-calling and MCP interaction formats. These design choices introduce diverse structural constraints on how models must issue tool calls, parse responses, and coordinate multi-step actions. Such diversity poses a significant challenge for a single model to generalize across different community-adopted IDEs/CLIs.

To systematically evaluate this crucial capability, we construct an in-house evaluation benchmark that explicitly measures model's adaptability to different real-world agentic coding scaffolds/IDEs. The benchmark consists of multiple prompt templates and tool-call schemas derived from representative IDE/CLI scaffolds, covering variations in system instructions, XML-variant/JSON-based tool call patterns.
For each question, we assess whether the model can follow the instructions in the system prompt and generate precisely formatted tool calls that satisfy the corresponding scaffold specifications.

\begin{table}[ht]
    \centering
    \begin{tabular}{c|cccccc}\toprule
         \textbf{Model}&  \textbf{Scaffold$_1$} &  \textbf{Scaffold$_2$} &  \textbf{Scaffold$_3$} & \textbf{Scaffold$_4$} & \textbf{Scaffold$_5$} & \textbf{Avg.} \\ 
         \midrule
         GPT-5-2 & 84.0 & 14.0 & 41.8 & 29.8 & 77.0 & 49.3  \\
         Claude-sonnet-4-5 &  86.8 & 61.0 & 100.0 & 88.0 & 91.0 & 85.4 \\
         Gemini-3-pro &  92.4 & 57.0 & 98.0 & 93.5  & 94.0 & 87.0 \\ \midrule
         Deepseek-v3.2 & 98.0 & 87.0 & 100.0 & 92.5 & 91.0 & 93.7 \\
         GLM-4.6 & 85.8 & 86.0 & 100.0 & 82.9 & 24.0 & 75.7 \\
         GLM-4.7 & 91.0 & 64.0 & 100.0 & 94.7  & 0.0 & 69.9\\
         MiniMax-M2.1  & 68.3 & 48.0 & 93.8 & 86.2 &  90.0 & 77.3\\
         Kimi-K2 & 59.0 & 59.0 & 100.0 & 57.4 & 81.0 & 71.3\\
         Kimi-K2-thinking & 71.5 & 70.0 & 91.8 & 83.0 & 80.0 & 79.3\\
         \midrule
         Qwen3-Coder-Next  & 98.0 & 83.0 & 98.0 & 91.5 & 93.0 & 92.7  \\
         \bottomrule
    \end{tabular}
    \caption{Template Following generalization accuracy across community-adopted IDE/CLI Environments.}
    \label{tab:template_following_bench}
\end{table}

As shown in Table~\ref{tab:template_following_bench}, while some models perform well on specific templates, their performance varies substantially across different IDE/CLI environments, indicating sensitivity to particular formatting conventions. In contrast, our model is able to follow consistently well on all five environments, demonstrating robust generalization to diverse prompt templates and tool-call schemas in real-world coding environments.

\subsubsection{Single-turn Question Answering Expert}
To further improve reasoning and complex coding ability, we apply reinforcement learning (RL) in execution-verifiable domains, focusing on single-turn coding tasks and complex instruction-following scenarios. In our overall RL framework, we consider two complementary regimes: single-turn RL, where correctness can be directly verified through execution (e.g., unit tests), and multi-turn agentic RL, where the model must interact with an environment over multiple steps. This section focuses on the single-turn RL setting, which primarily targets code reasoning tasks (e.g., competitive programming) and complex instruction-following tasks that can be evaluated via execution.

\begin{figure}[t]
    \centering
    \includegraphics[width=1.0\textwidth]{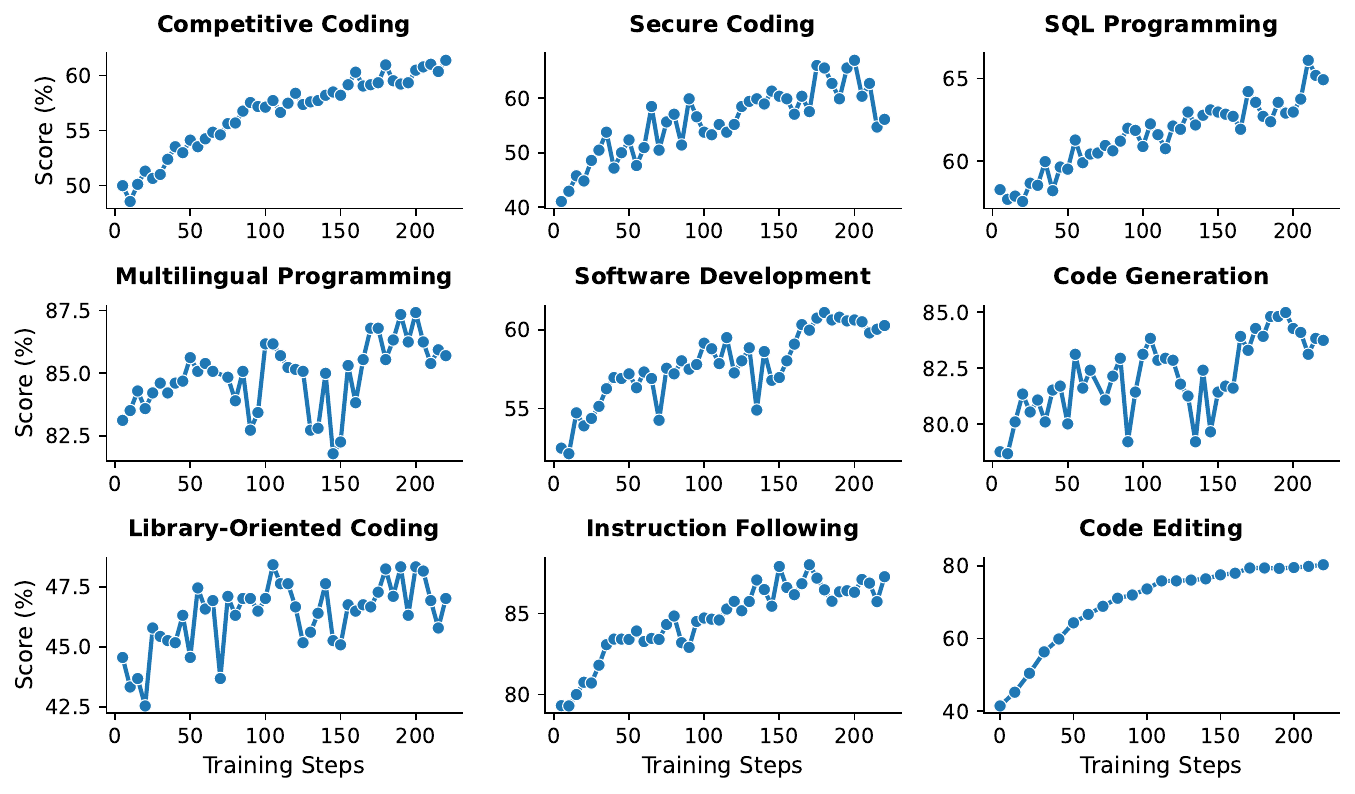}
    \caption{Performance trends of various coding sub-capabilities throughout the single-turn RL steps, measured by our in-house benchmarks.}
    \label{fig:single-rl}
\end{figure}

For single-turn RL, instead of focusing exclusively on competitive programming as in much prior work, we argue that most coding tasks are naturally well-suited for execution-driven reinforcement learning. The key reason is that code correctness can be directly verified by running it against unit tests, which provides a reliable and scalable learning signal. Following this view, we extend code RL training to a broader set of realistic coding tasks, aiming to better exploit the potential of large-scale RL beyond competitive programming.

To this end, we synthesize tasks that cover a broader spectrum of programming competencies. In particular, many tasks require library usage, including calling standard or third-party APIs, handling I/O and data formats, and composing existing utilities. These requirements more closely match real-world development scenarios than purely algorithmic problems. We further extend the task suite to multiple programming languages. This multilingual setup encourages the model to learn language-specific idioms, tooling constraints, and semantic differences, such as type systems, standard libraries, error handling behavior, and runtime characteristics, instead of overfitting to a single-language distribution. We also consider vulnerability-prone coding scenarios, where the model is asked to generate secure code snippets and repair the vulnerabilities. For both tasks, we ensure that the functional correctness and security are properly covered.

After scaling the task collection, we automatically synthesize corresponding unit tests for each instance. To obtain reliable unit tests without human annotation, we generate multiple candidate unit tests using internal models and retain the tests that achieve the highest consensus under majority voting across independently generated solutions. These unit tests are then used to drive RL through execution-based rewards. As shown in \Cref{fig:single-rl}, scaling RL task diversity leads to consistent improvements across multiple coding sub-capabilities.




\subsubsection{Software Engineering Expert}\label{sec:swe}

Real-world software engineering tasks require models to reason over large codebases, interact with tools and execution environments, and operate reliably across long interaction horizons. To address these challenges, we train a Software Engineering expert specialized for multi-step, environment-interactive coding tasks.

\paragraph{Data.} RL queries are derived from real-world software engineering tasks, including open-source datasets \citep{pan2025training,swe-rebench} and our automatically constructed repository environments (\cref{sec:task_synthesis}). 


To prevent information leakage between training stages, SFT and RL prompts are fully disjoint. In addition, we estimate the pass-rate distribution of each training instance and filter out both overly easy examples and noisy failure cases. This allows RL training to focus on informative failures that provide stronger learning signals.

\paragraph{Reward Shaping.} In multi-turn RL rollouts, the model interacts with the environment through tool calls across multiple steps to solve software engineering tasks. Trajectory-level rewards are assigned based on final task completion. However, correct final outcomes do not necessarily imply high-quality intermediate reasoning or tool usage \citep{shum2025swerm}. To address this, we introduce additional trajectory-level and token-level penalties.

First, we apply an unfinished trajectory penalty. When the number of interaction turns exceeds a predefined maximum, the trajectory reward is penalized to discourage excessively long rollouts and failure to terminate.

Second, we apply a turn-level tool-format penalty. At each interaction step, we perform rule-based validation of tool-call format correctness. During optimization, tokens associated with invalid tool calls receive token-level penalties, preventing the model from learning malformed tool invocation patterns.

\begin{figure}[t]
    \centering
    \includegraphics[width=1.0\textwidth]{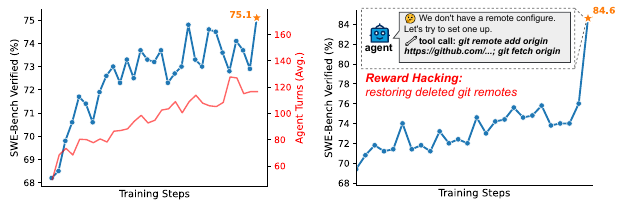}
    \vspace{-1em}
    \caption{\textbf{Left}: SWE-bench Verified performance vs. RL steps with a reinforced reward-hacking blocker. We also found that a \textit{long-horizon} coding ability emerged in the model during RL training, pushing the average number of agent turns from 50 to 130. \textbf{Right}: Performance without the blocker. Even after removing \texttt{git} remotes and future commits, the agent autonomously learns to exploit various \texttt{git} commands to retrieve ground-truth information as model capability increases. To the best of our knowledge, this behavior has not previously been reported.}
    \label{fig:multi-rl}
\end{figure}



\paragraph{Reinforced Reward Hacking Blocker.}

Prior work has shown that GitHub-based environments may unintentionally leak future commit information\footnote{\url{https://github.com/SWE-bench/SWE-bench/pull/471}}, which agents can exploit to recover ground-truth fixes (e.g., via \texttt{git log -{}-all}). To mitigate this, we adopt standard protections including removing remotes, branches, and tags.

During later RL stages, however, many new ways of reward hacking emerge. Agents attempt to reconnect local repositories to GitHub using commands such as \texttt{git remote add}, or retrieve commit history through \texttt{git clone}, \texttt{curl}, or similar tools, as illustrated in \Cref{fig:multi-rl}. Fully disabling network access is not reasonable, as agents require connectivity for legitimate operations such as environment setup, documentation retrieval, or installing additional packages.

To address this, we introduce a heuristic blocking rule. Any tool call containing both a repository link (e.g., \texttt{github.com/\{repo\}}) and network-access keywords (e.g., \texttt{git}, \texttt{curl}, \texttt{wget}) is blocked, and the agent receives explicit feedback indicating the prohibited action. With our improved blocker, our manual inspection of trajectories confirms that reward-hacking behaviors are effectively eliminated.

\subsubsection{Expert Distillation}
Finally, we perform expert distillation to consolidate capabilities from multiple domain experts into a single unified deployment model. Concretely, we distill knowledge from domain-specialized experts, including Web Development, User Experience, Single-turn RL, and Software Engineering experts, into the SFT model.

Through distillation, the unified model inherits the strengths of individual experts while preserving the strong instruction following capability of the base SFT model. This enables practical deployment in real-world agentic coding scenarios, where a single model must handle diverse tasks spanning multiple domains without relying on expert routing or multi-model orchestration.

\section{Experiments}
\label{sec:experiments}
\begin{table}[t]
\centering
\small
\setlength{\tabcolsep}{6pt}
\begin{tabular}{l c ccc}
\toprule
\multirow{2}{*}{Model} & \multirow{2}{*}{Size} &
\multicolumn{3}{c}{SWE-Bench Verified} \\
\cmidrule(lr){3-5}
& & SWE-Agent & MiniSWE-Agent & OpenHands \\
\midrule
\textit{Proprietary Models} & & \\
Claude-Opus-4.5    & ?         & 78.2 & 77.8 & 79.0 \\
Claude-Sonnet-4.5  & ?         & 76.0 & 68.4 & 74.6 \\
\midrule
\textit{Open-source Models} & & \\
DeepSeek-V3.2      & 671A37    & 70.2 & 67.2 & 72.6 \\
GLM-4.7             & 358A32    & 74.2 & 70.4 & 70.6 \\
MiniMax-M2.1         & 230A10    & 74.8 & 70.4 & 71.0 \\
Kimi-K2.5           & 1000A32 & 73.2 & 70.8 & --   \\
\textbf{Qwen3-Coder-Next}   & 80A3      & 70.6 & 71.1 & 71.3 \\
\bottomrule
\end{tabular}
\caption{SWE-Bench Verified. Dashes indicate we couldn't reliably obtain test results in such settings. The maximum number of agent turns was set to 300.}
\label{tab:swe-bench-verified}
\end{table}

\begin{table}[t]
\centering
\small
\setlength{\tabcolsep}{6pt}
\resizebox{\textwidth}{!}{%
\begin{tabular}{l c ccc cc}
\toprule
\multirow{2}{*}{Model} & \multirow{2}{*}{Size} &
\multicolumn{3}{c}{SWE-Bench Multilingual} &
\multicolumn{2}{c}{SWE-Bench-Pro} \\
\cmidrule(lr){3-5}\cmidrule(lr){6-7}
& & SWE-Agent & MiniSWE-Agent & OpenHands & SWE-Agent & MiniSWE-Agent \\
\midrule
\textit{Proprietary Models} & & \\
Claude-Opus-4.5    & ?         & 71.7 & 71.8 & 75.2 & 51.6 & 50.2 \\
Claude-Sonnet-4.5  & ?         & 67.2 & 61.3 & 66.0 & 50.5 & 43.0 \\
\midrule
\textit{Open-source Models} & & \\
DeepSeek-V3.2      & 671A37    & 62.3 & 55.5 & 61.8 & 46.0 & 32.4 \\
GLM-4.7             & 358A32    & 63.7 & 61.8 & 60.8 & 45.1 & 39.4 \\
MiniMax-M2.1         & 230A10    & 66.2 & 62.5 & 67.5 & 40.8 & 39.1 \\
Kimi-K2.5           & 1000A32 & 63.7 & 62.3 & --   & 47.3 & 42.8 \\
\textbf{Qwen3-Coder-Next}   & 80A3      & 62.8 & 56.2 & 64.3 & 42.7 & 38.7 \\
\bottomrule
\end{tabular}
}
\caption{SWE-Bench Multilingual and SWE-Bench Pro. Dashes indicate we couldn't reliably obtain test results in such settings. The maximum number of agent turns was set to 300.}
\label{tab:swe-bench-multi-pro}
\end{table}

\subsection{Agentic Evaluation}

\paragraph{Setting.} We evaluate \model on three SWE benchmarks, SWE-Bench Verified~\citep{swe-bench}, SWE-Bench Multilingual~\citep{swe-smith}, and SWE-Bench Pro~\citep{deng2025swebenchproaiagents}, and a command-line interface (CLI) task, TerminalBench 2.0~\citep{merrill2026terminalbenchbenchmarkingagentshard}. To ensure a fair comparison, we replicated all baselines on each scaffold and adopted standard hacking-free mechanisms, including the removal of remotes, branches, and tags, to prevent the agent from accessing future commit information, as we introduce in \Cref{sec:swe}. The baselines include two proprietary models, Claude-Opus-4.5~\citep{claude45opus} and Claude-Sonnet-4.5~\citep{claude-sonnet-4-5}, as well as four top-tier open-source models, DeepSeek-V3.2~\citep{deepseekv32}, GLM-4.7~\citep{glm-4-7}, MiniMax-M2.1~\citep{minimax-m2.1}, and Kimi-K2.5~\citep{kimi-k25}. The maximum number of agent turns was set to 300.

\paragraph{SWE Coding Tasks.} \Cref{tab:swe-bench-verified} reports results on SWE-Bench Verified across three agent scaffolds. \model achieves strong and consistent performance across SWE-Agent, MiniSWE-Agent, and OpenHands, remaining competitive with substantially larger frontier and open-weight models. Specifically, it achieves scores of 70.6\% with SWE-Agent, 71.1\% with MiniSWE-Agent, and 71.3\% with OpenHands. A key highlight is the model's exceptional efficiency. Despite its relatively compact size of 80A3, \model achieves results that are on par with, and in some cases outperform, much larger models.

\Cref{tab:swe-bench-multi-pro} presents results on SWE-Bench Multilingual, which evaluates repository-level bug fixing across multiple programming languages, as well as the more challenging SWE-Bench Pro benchmark. On SWE-Bench-Pro, which emphasizes longer-horizon software engineering tasks, \model remains competitive with significantly larger open-weight models, highlighting a favorable efficiency–performance trade-off under multilingual and high-difficulty evaluation conditions.



\begin{table}[t]
\centering
\small
\setlength{\tabcolsep}{6pt}
\begin{tabular}{l c cccc}
\toprule
\multirow{2}{*}{Model} & \multirow{2}{*}{Size} & \multicolumn{4}{c}{TerminalBench 2.0} \\
\cmidrule(lr){3-6}
& & Terminus2-xml & Terminus2-json & ClaudeCode & QwenCode \\
\midrule
\textit{Proprietary Models} & & \\
Claude-Opus-4.5    & ?         & 58.4 & 57.3 & 53.9 & 51.7 \\
Claude-Sonnet-4.5  & ?         & 51.7 & 51.7 & 41.6 & 37.1 \\
\midrule
\textit{Open-source Models} & & \\
DeepSeek-V3.2      & 671A37    & 34.8 & 39.3 & --   & --   \\
GLM-4.7             & 358A32    & 44.9 & 37.1 & --   & 31.5 \\
MiniMax-M2.1         & 230A10    & --   & 32.6 & 42.7 & 39.3 \\
Kimi-K2.5           & 1000A32   & 38.8   & 49.4   & \phantom{0}9.0   & 27.5   \\
\textbf{Qwen3-Coder-Next}   & 80A3      & 34.2 & 36.2 & 30.9 & 25.8 \\
\bottomrule
\end{tabular}
\caption{Terminal-Bench 2.0 results. Dashes indicate we couldn't reliably obtain test results in such settings.}
\label{tab:tb}
\end{table}

\paragraph{Command-line Interface tasks.} \Cref{tab:tb} shows the results on Terminal Bench 2.0, \model demonstrates consistent performance across multiple TerminalBench 2.0 environments, including XML- and JSON-based tool schemas as well as different agent scaffolds. These results suggest that the model's training on diverse tool-call formats and agentic workflows translates effectively to real-world interactive coding settings. While there is clear room for improvement in this area, \model establishes a strong and efficient foundation for complex tool-use tasks.

\begin{table}[t]
\centering
\small
\begin{tabular}{@{}lcccccc@{}}
\toprule
Model & EvalPlus & MultiPL-E & CRUXEval & LiveCodeBench (v6) & OJBench & Codeforces \\
\midrule
Qwen3-Coder-480B-A35B & 86.66 & 88.00 & 92.13 & 44.93 & 14.98 & 1800 \\
Qwen3-Next            & 89.00 & 89.00 & 94.81 & 51.79 & 20.04 & 1875 \\
\textbf{Qwen3-Coder-Next}      & 86.56 & 88.23 & 95.88 & 58.93 & 23.01 & 2100 \\
\bottomrule
\end{tabular}
\vspace{-0.75em}
\caption{Results on function-level coding and competitive-programming benchmarks.}
\label{tab:other-coding-1}
\end{table}

\begin{table}[t]
\centering
\small
\begin{tabular}{@{}lccccc@{}}
\toprule
Model & FullStackBench-en & FullStackBench-zh & Spider & BIRD-SQL & Aider-Polyglot \\
\midrule
Qwen3-Coder-480B-A35B & 62.54 & 63.07 & 85.98 & 64.15 & 60.40 \\
Qwen3-Next            & 62.30 & 59.22 & 82.50 & 66.62 & 52.90 \\
\textbf{Qwen3-Coder-Next}      & 60.58 & 57.38 & 83.66 & 63.56 & 66.20 \\
\bottomrule
\end{tabular}
\vspace{-0.75em}

\caption{Results on full-stack development, text-to-SQL, and multilingual code editing benchmarks.}
\label{tab:other-coding-2}
\end{table}

\subsection{Other Coding Tasks}

We evaluate \model on a broader suite of coding benchmarks covering unit-test style evaluation \citep{evalplus,multiple}, code reasoning \citep{gu2024cruxeval}, competitive-programming style tasks \citep{livecodebench,wang2025ojbenchcompetitionlevelcode}, full-stack development \citep{liu2024fullstackbenchevaluatingllms}, and structured querying \citep{yu-etal-2018-spider,li2024can}, as well as multi-language code editing\footnote{\url{aider.chat/docs/leaderboards/edit.html}}. Results are presented in \Cref{tab:other-coding-1} and \Cref{tab:other-coding-2}, where we compare \model to Qwen3-Coder-480B-A35B, our last flagship coder model, and Qwen3-Next, a general model. Overall, these results show that \model offers a favorable trade-off, with consistent gains on more difficult competitive-programming and reasoning benchmarks while maintaining solid performance across full-stack and data-centric coding tasks.

\subsection{General Tasks}
\begin{table}[h!]
\centering
\small
\setlength{\tabcolsep}{6pt}
\begin{tabular}{@{}lccccc@{}}
\toprule
Model & MMLU & MMLU-Redux & MMLU-Pro & GPQA & SuperGPQA \\
\midrule
Qwen3-Next & 87.87 & 91.14 & 80.89 & 73.54 & 58.70 \\
\textbf{\model} & 87.73 & 91.18 & 80.52 & 74.49 & 57.45 \\
\bottomrule
\end{tabular}
\vspace{-0.75em}

\caption{Results on general knowledge and reasoning benchmarks.}
\label{tab:general-knowledge}
\end{table}

\begin{table}[h!]
\centering
\small
\setlength{\tabcolsep}{6pt}
\begin{tabular}{@{}lcccc@{}}
\toprule
Model & HMMT25 Feb & HMMT25 Nov & AIME24 & AIME25 \\
\midrule
Qwen3-Next  & 54.27 & 68.07 & 82.92 & 69.64 \\
\textbf{\model}  & 70.21 & 75.57 & 89.01 & 83.07 \\
\bottomrule
\end{tabular}
\vspace{-0.75em}

\caption{Results on competitive math benchmarks.}
\label{tab:math}
\end{table}

We compare \model to Qwen3-Next on general knowledge \citep{mmlu,mmlupro,mmluredux} and reasoning \citep{gpqa,supergpqa} benchmarks, and results are shown in \Cref{tab:general-knowledge}. Despite being a model specialized for coding, its general capability remains strong. \model is competitive with Qwen-Next, slightly improving on MMLU-Redux and GPQA, while remaining close on MMLU, MMLU-Pro, and SuperGPQA.

We also compare \model to Qwen3-Next on competitive math benchmarks \citep{aime,balunovic2025matharena}, and results are presented in \Cref{tab:math}. \model substantially outperforms the baseline across all benchmarks, with large gains on HMMT25 Feb and AIME25, and clear improvements on HMMT25 Nov and AIME24. These results indicate that strong code reasoning capabilities can be transferred to math reasoning capabilities.

\section{Conclusion, Limitation, and Future Work}

In this work, we introduced \model, a coding model designed for real-world agentic software development tasks. Built on a hybrid mixture-of-experts architecture with $80$ billion total parameters and only $3$ billion active parameters per forward pass, \model is designed to balance strong reasoning and coding capability with practical inference efficiency. By scaling agentic training through large-scale synthesis of executable coding tasks and learning from execution feedback, we significantly improve tool-use robustness, long-context coding ability, and multi-domain coding performance.

Despite these advances, we acknowledge several limitations compared with frontier proprietary models such as Claude Opus 4.5. \model is designed with a significantly smaller active compute footprint and lower total training compute, which enables efficient deployment but introduces capability trade-offs. While the model demonstrates strong instruction-following and solid coding fundamentals, there remains a gap in solving highly complex, large-scale software engineering tasks, which we plan to address by scaling exposure to harder and more realistic software projects during pre-training. In addition, for some complex tasks, the model may require more interaction turns to reach a correct solution, motivating future work on improving reasoning efficiency through reinforcement learning and better long-horizon planning. Furthermore, frontend and UI-related capability remains an area for improvement. Finally, we aim to explore agentic and real-world cybersecurity tasks for the future models, such as vulnerability exploitation and capture-the-flag competitions. such as  We plan to address this by integrating visual capability into future agent models, enabling the model to directly evaluate rendered outputs and interactive behavior.
\clearpage
\section{Authors}

\textbf{Core contributors:} Ruisheng Cao, Mouxiang Chen, Jiawei Chen, Zeyu Cui, Yunlong Feng, Binyuan Hui, Yuheng Jing, Kaixin Li, Mingze Li, Junyang Lin, Zeyao Ma, Kashun Shum, Xuwu Wang, Jinxi Wei, Jiaxi Yang, Jiajun Zhang, Lei Zhang, Zongmeng Zhang, Wenting Zhao, Fan Zhou

\textbf{Contributors:} Tianyi Bai, Keqin Bao, Chen Cheng, Yizhong Cao, Xiaodong Deng, Shichun Feng, Hao Ge, Fei Huang, Yukai Huang, Fangyu Lei, Xiaochuan Li, Ziyang Li, Dayiheng Liu, Yukun Liu, Yuqiong Liu, Zhongwei Liu, Chen Liang, Dunjie Lu, Rui Men, Yuandong Ni, Xingzhang Ren, Yang Su, Jianhong Tu, Junli Wang, Yongxing Wu, Chencan Wu, Tianbao Xie, Yiheng Xu, Mingfeng Xue, An Yang, Kexin Yang, Zhiyu Yin, Beichen Zhang, Yi Zhang, Jianwei Zhang, Kai Zhang, Bo Zheng, Jingren Zhou, Terry Yue Zhuo

Authors are listed alphabetically by their last names.

\bibliography{biblio}
\bibliographystyle{colm2024_conference}

\clearpage
\appendix
\section{Appendix}
\label{sec:appendix}

\subsection{Data Statistics for Synthesized Tasks}
\label{app:data_statistics_for_cpt}

\paragraph{Building repository-level environments with agents.} In \Cref{tab:real-world-pr}, we present detailed data statistics for all repositories built from real-world GitHub pull requests.


\begin{table}[h]
\centering
\small
\caption{Data statistics for real-world repository instances.}
\label{tab:real-world-pr}
\begin{tabular}{lrrrr}
\toprule
\textbf{Language} & \textbf{Instances} & \textbf{Repos} & \textbf{Inst/Repo (Avg)} & \textbf{Avg Eval Lines} \\
\midrule
Python & 202,302 & 13,098 & 15.45 & 25.01 \\
Javascript / Typescript & 175,660 & 11,604 & 15.14 & 27.41 \\
Go & 121,062 & 5,554 & 21.80 & 28.87 \\
Java & 86,105 & 4,700 & 18.32 & 24.75 \\
Rust & 74,180 & 4,445 & 16.69 & 19.31 \\
C / C++ & 37,228 & 3,405 & 10.93 & 45.78 \\
C\# & 24,387 & 1,929 & 12.64 & 31.84 \\
Others & 86,769 & 8,225 & 10.55 & 38.89 \\
\midrule
\textbf{\# Total} & 807,693 & 52,960 & 15.25 & 28.21 \\
\bottomrule
\end{tabular}
\end{table}

\paragraph{Synthesizing issues.} In \Cref{tab:agent_env_3_data_statistics_for_cpt}, we present detailed data statistics for all repositories collected from open-source projects and synthesized bugs with different bug sampling strategies. On average, we succeed to sample $169.7$ bugs (or tasks) for each code repository.

\begin{table}[htbp]
  \centering
  \caption{Data statistics for generated task instances with method workflow-based pipelined bug synthesis.}
  \label{tab:agent_env_3_data_statistics_for_cpt}
  \resizebox{1.0\textwidth}{!}{%
    \begin{tabular}{cc|ccc|cccc|c}
    \toprule
    \multirow{2}{*}{\textbf{Dataset}} & 
    \multicolumn{1}{c|}{\multirow{2}{*}{\textbf{PL}}} & 
    \multicolumn{1}{c}{\multirow{2}{*}{\makecell{\textbf{\# Repos} \\ \textbf{(raw)}}}} & 
    \multicolumn{1}{c}{\multirow{2}{*}{\makecell{\textbf{\# Repos} \\ \textbf{(cleaned)}}}} & 
    \multicolumn{1}{c|}{\multirow{2}{*}{\makecell{\textbf{\# Repos} \\ \textbf{(used)}}}} & 
    \multicolumn{4}{c|}{\textbf{Bug Sampling Strategy}} & 
    \multirow{2}{*}{\textbf{\# Total}} \\
    \cmidrule{6-9}
    & & & & & 
    \texttt{lm\_rewrite} & 
    \texttt{lm\_modify} & 
    \texttt{func\_pm} & 
    \texttt{others} & \\
    \midrule
    SWE-smith & Python & 134 & 134 & 130 & 10,980 & 23,120 & 28,467 & 11,436 & 74,003 \\
    SWE-Flow & Python & 2,203 & 2,203 & 1,987 & 81,963 & 119,892 & 182,686 & - & 384,541 \\
    SWE-rebench & Python & 3,468 & 2,912 & 2,727 & 48,660 & 80,246 & 244,219 & - & 373,125 \\
    SWE-smith-multi & Multilingual & 133 & 133 & 118 & 2,448 & 6,749 & 3,401 & 1,065 & 13,663 \\
    Multi-SWE-RL & Multilingual & 74 & 74 & 57 & 1,399 & 3,362 & 1,805 & - & 6,566 \\
    \midrule
    \multicolumn{2}{c|}{\textbf{Total}} & 
    6,012 & 5,456 & \textbf{5,019} & 
    145,450 & 233,369 & 460,578 & 12,501 & 
    \textbf{851,898} \\
    \bottomrule
    \end{tabular}%
  }%
\end{table}

\subsection{The Checklist of Tool Chat Templates for Scaling}
\label{app:tool_chat_templates}
In Table \ref{tab:tool_chat_templates}, we list all $21$ tool chat templates we used for scaling. These templates, originating from open-source models and agent scaffolds, differ mainly from each other in the formats of both tool definition and tool call formats. Note that, {\tt qwen3\_xml\_mixed} is a variant of {\tt qwen3\_coder}, where the tool definition is structured as JSON lines. \texttt{hermes} is a generic JSON-format function calling templates. \texttt{harmony\_json} and \texttt{harmony\_xml} are adapted from the original harmony format in {\tt gpt-oss}~\citep{gpt-oss}. Template \texttt{xml\_cline} and \texttt{xml\_aone} are summarized from agent scaffolds Cline~\footnote{Cline: \url{https://marketplace.visualstudio.com/items?itemName=saoudrizwan.claude-dev}.} and Aone Copilot~\footnote{Aone Copilot: \url{https://marketplace.visualstudio.com/items?itemName=Aone.aone-copilot},} respectively.
\begin{table}[htbp]
  \centering
  \caption{The checklist of all used tool chat templates for scaling.}

  \resizebox{1.0\textwidth}{!}{
    \begin{tabular}{c|c|c|c}
    \toprule
    \textbf{Name} & \textbf{Model / Agent} & \textbf{Tool Definition Format} & \textbf{Tool Calls Format} \\
    \midrule
    \texttt{qwen3\_coder} & Qwen-Coder-Next (this work) & XML   & XML   \\
    \texttt{qwen3\_xml\_mixed} & Qwen-Coder-Next (this work) & JSON  & XML \\
    \texttt{deepseekr1} & DeepSeek-R1~\citep{r1} & JSON  & Mixed XML+JSON \\
    \texttt{deepseekv3} & DeepSeek-V3~\citep{deepseekv3} & JSON  & Mixed XML+JSON \\
    \texttt{deepseekv31} & DeepSeek-V3.1~\citep{deepseekv3} & Text  & Mixed XML+JSON  \\
    \texttt{deepseekv32} & DeepSeek-V3.2~\citep{deepseekv32} & JSON  & XML \\
    \texttt{glm46} & GLM-4.6~\citep{glm46} & JSON  & XML \\
    \texttt{minimax\_m1} & MiniMax-M1~\citep{minimax_m1} & JSON  & JSON  \\
    \texttt{minimax\_m2} & MiniMax-M2~\citep{minimax_m2} & XML   & XML  \\
    \texttt{kimik2} & Kimi-K2~\citep{kimik2} & JSON  & Mixed XML+JSON \\
    \texttt{hermes} & -     & JSON  & JSON  \\
    \texttt{qwen25\_coder} & Qwen2.5-Coder~\citep{qwen2.5coder} & JSON  & JSON \\
    \texttt{harmony\_json} & gpt-oss~\citep{gpt-oss} & TypeScript & JSON  \\
    \texttt{harmony\_xml} & gpt-oss~\citep{gpt-oss} & TypeScript & XML\\
    \texttt{llama4\_pythonic} & Llama4~\citep{llama4} & JSON  & Python \\
    \texttt{toolace} & ToolACE-8B~\citep{toolace} & JSON  & Python  \\
    \texttt{mistral3} & Mistral-Large-3~\citep{mistral3} & JSON  & Text + JSON \\
    \texttt{xlam\_qwen} & xLAM-2~\citep{xlam} & JSON  & JSON  \\
    \texttt{xml\_cline} & Cline Agent & JSON & XML \\
    \texttt{xml\_aone} & Aone Copilot  & JSON & XML  \\
    \bottomrule
    \end{tabular}%
  }
  \label{tab:tool_chat_templates}%
\end{table}%

\subsection{Detailed Implementation on Best-Fit-Packing}
\label{app:best_fit_packing}
The traditional sample packing strategy during pre-training first 
concatenates all documents into a single flattened text sequence, 
inserting a special separator token (e.g., "<|endoftext|>") between 
documents. It then splits this entire sequence into fixed-length chunks according to the model's context size to construct training samples. Though efficient due to zero padding, it inevitably incurs the notorious context hallucination problem~\citep{hallucination}. This problem is particularly pronounced in multi-turn agent interaction tasks involving tool calls, where all tool definitions and their calling formats are typically defined only at the beginning of trajectories. Random document chunking prevents models from learning and adhering to the strict formatting requirements of tool invocations. To this end, we re-implement best-fit-packing algorithm~\citep{best-fit-packing} with C++ in the Megatron framework~\citep{megatron}, and compare it with two other variants quantitively.

\paragraph{Fragmentation Rate and Padding Rate}
To formally analyze the impact of sample packing strategies, we define two metrics, fragmentation rate and padding rate, to report 1) the proportion of fragmented documents over the total number of training documents, and 2) the number of padding tokens (which are not trained upon) against the total number of training tokens, respectively. Mathematically,
\begin{align*}
    \text{fragmentation rate}&=\frac{\text{\# of documents that are fragmented}}{\text{\# of all documents}},\\
    \text{padding rate}&=\frac{\text{\# of padding tokens}}{\text{\# of all training tokens}}.
\end{align*}

\subsubsection{Two Variants of Sample Packing Strategy}
\begin{figure}[htbp]
    \centering
    \includegraphics[width=0.98\textwidth]{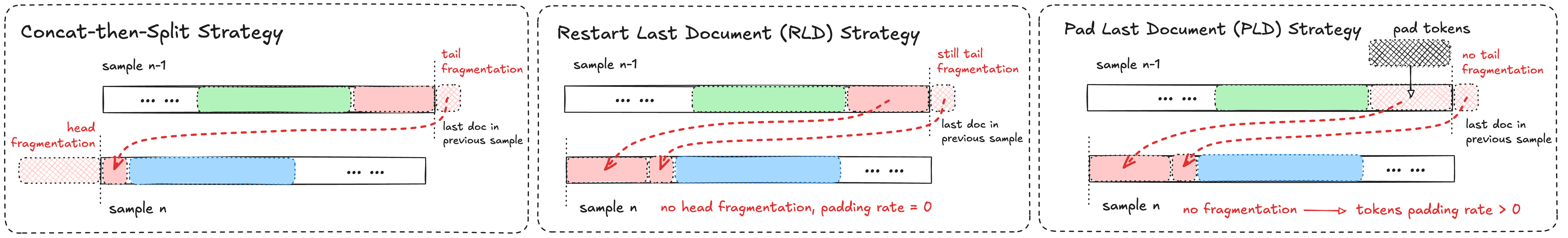}
    \caption{Two simple variants of the traditional concat-then-split sample packing strategy: restart last document~(RLD) and pad last document~(PLD).}
    \label{fig:bfp_variants}
\end{figure}

For compatibility with the data loader in Megatron, we propose two simple variants with minimal modifications of the original concat-then-split packing strategy~(see Figure~\ref{fig:bfp_variants}).

\paragraph{Restart Last Document~(RLD)} Aiming to resolve the head fragmentation problem, the last fragmented document in the previous chunk is forced to restart in the current sample. That is, each training sample will start with the beginning of one document, completely eliminating the fragmentation at the head side. Another benefit is that the padding rate is still zero, which implies optimal training efficiency. However, the tail side fragmentation is preserved. And this strategy evidently introduces a reweighting of tokens: tokens at the beginning of long documents, due to their higher likelihood of being reused (or restarted), are implicitly assigned greater weight during training.

\paragraph{Pad Last Document~(PLD)} Based on RLD, in order to prevent the aforementioned tail-side truncation and imbalanced token weight re-distribution, we fill in the last fragmented document of the previous training sample with padding tokens. The gradients of these padding tokens are masked during pre-training to prevent unreasonable back-propagation. Although this scheme completely eliminates the fragmentation issue, it sacrifices training efficiency for a significant ratio of padding tokens. For fair comparison, we need to scale up the total size of training tokens in proportion. Formally, the updated token size is calculated by $$\text{\# of training tokens with PLD}=\text{\# of training tokens}\times \frac{1}{1-\text{padding\_rate}}.$$

\subsubsection{How to Tackle Long Documents with Best-Fit-Packing}
The original BFP algorithm~\citep{best-fit-packing} treats sample packing as a classic bin packing problem, i.e., for each incoming document, it seeks a bin~(or sample) with sufficient remaining capacity to accommodate it entirely. The prerequisite is that all documents are shorter than the model input length. When the maximum context length cannot individually accommodate an extremely long document, we propose the following three techniques to handle it:
\begin{itemize}[left=4pt]
    \item \textbf{Split:} It pre-splits extremely long documents into chunks with sizes equal to the model input length. The last chunk, which may be shorter, is preserved as is.
    \item \textbf{Slide:} It pre-splits extremely long documents using a sliding window with overlap. The last chunk is merged with the previous one or extended backward to maintain the target model length, trying to infer based on more context.
    \item \textbf{Drop:} It aggressively drops those extremely long documents before running BFP.
\end{itemize}

\subsubsection{Ablation Study on Sample Packing Strategy}
For efficient evaluation without containerized environment or agent interaction, we experiment with an agentless framework on SWE-bench~\citep{swe-bench}. This agentless pipeline is adapted from \cite{agentless}, which splits the entire issue-solving task into two steps: 1) bug localization, and 2) patch prediction. As for bug localization, we prepare three settings to further simplify the evaluation, where the bug location comes from 1) model prediction, 2) the ground truth bug snippet, and 3) the ground truth file containing the bug. In this way, the trained model only needs to predict the target patch, based on different location implications. In \Cref{tab:bfp_experiments}, we report two metrics for each setting: 1) the similarity between the predicted patch and the oracle patch, and 2) the ratio of empty patch.

\begin{table}[htbp]
  \centering
  \caption{Ablation study on different sample packing strategies and how to tackle long documents exceeding model input length. Model Loc: use model predicted position for bug localization; GT Loc: use ground truth bug location; GT File: use the complete file containing bug or target solution. $\uparrow$ means higher is better, while $\downarrow$ means lower is better.}
  \resizebox{1.0\textwidth}{!}{
    \begin{tabular}{r|c|cc|cc|cc|cc|cc}
    \hline

    \hline
    \multicolumn{1}{c|}{\multirow{2}{*}{\textbf{Strategy}}} & \multirow{2}{*}{\textbf{Tokens (B)}} & \multicolumn{1}{c}{\multirow{2}{*}{\makecell{\textbf{Fragm.} \\ \textbf{Rate (\%)}}}} & \multicolumn{1}{c|}{\multirow{2}{*}{\makecell{\textbf{Padding} \\ \textbf{Rate (\%)}}}} & \multicolumn{2}{c|}{\textbf{Model Loc (\%)}} & \multicolumn{2}{c|}{\textbf{GT Loc (\%)}} & \multicolumn{2}{c|}{\textbf{GT File (\%)}} & \multicolumn{2}{c}{\textbf{AVG (\%)}} \\
\cline{5-12}          &       &       &       & \textbf{sim $\uparrow$} & \textbf{empty $\downarrow$} & \textbf{sim $\uparrow$} & \textbf{empty $\downarrow$} & \textbf{sim $\uparrow$} & \textbf{empty $\downarrow$} & \textbf{sim $\uparrow$} & \textbf{empty $\downarrow$} \\
    \hline
    \hline
    \multicolumn{1}{l|}{concat-then-split} & 73    & 30.2  & 0.00  & 16.61  & 30.61  & 24.02  & 26.60  & 9.41  & 59.60  & 16.68  & 38.94 \\
    \multicolumn{1}{l|}{restart-last-document} & 73    & 17.8  & 0.00  & \textbf{17.92 } & 25.92  & 23.51  & 27.00  & 10.28  & 54.60  & 17.24  & 35.84  \\
    \multicolumn{1}{l|}{pad-last-document} & 89    & 0.0   & 17.55  & 17.24  & \textbf{25.31 } & 22.99  & 25.60  & 10.35  & \textbf{52.40} & 16.86  & \textbf{34.44 } \\
    \multicolumn{1}{l|}{best-fit-packing} & 73    & 0.0   & 0.01  & 17.22  & 28.37  & \textbf{25.76 } & \textbf{23.60 } & \textbf{10.47 } & 56.80  & \textbf{17.82 } & 36.26 \\
    \hline
    \multicolumn{1}{l|}{best-fit-packing w/ split} & 73    & 0.0   & 0.01  & 17.95  & 22.04  & 27.46  & \textbf{16.80} & 15.10  & 36.20  & 20.17  & 25.01  \\
    w/ slide & 73    & 0.0   & 0.01  & \textbf{18.47 } & \textbf{20.41 } & 27.50  & 18.60  & 14.49  & 36.40  & 20.15  & 25.14  \\
    w/ drop & 73    & 0.0   & 0.01  & 17.85  & 21.43  & \textbf{29.53 } & 18.00  & \textbf{15.14 } & \textbf{33.60 } & \textbf{20.84 } & \textbf{24.34 } \\
    \hline

    \hline
    \end{tabular}%
    }
  \label{tab:bfp_experiments}%
\end{table}%

The results above reveal three key insights: 1) eliminating fragmentation steadily improves performance. Best-fit-packing outperforms the traditional concat-then-split strategy in both patch similarity and empty rate. 2) BFP is more token-efficient than padding, achieving better results ($17.82\%$ vs $16.86\%$) with $22\%$ fewer tokens. 3) handling extremely long documents is also non-negligible. Augmenting BFP with the ``drop'' strategy yields the best overall performance ($20.84\%$ similarity, $24.34\%$ empty rate). This suggests that, rather than fancy algorithmic improvements, the best solution for handling long contexts may be to directly extend the model context length. We adopt the ``split'' strategy to handle extremely long documents throughout the main experiments.

\subsection{Experiments on Cybersecurity}
We view cybersecurity as a frontier direction for existing models, where they have yet to achieve human-expert level performance even on non-agentic tasks such as Cyber Threat Intelligence (CTI) analysis, vulnerability detection, and secure coding. To provide insight into future capabilities, we present the first comparative evaluation of \model against other frontier models in this domain.

\begin{table}[ht]
\centering
\begin{tabular}{lcccccc}
\toprule
Model & CKT (Acc) & ATE (Acc) & RCM (Acc) & RMS (F1) & VSP (Acc) & TAA (Acc) \\
\midrule
Claude-Opus-4-5 & 94.33 & 85.50 & 71.25 & 68.39 & 53.88 & 29.00 \\
Claude-sonnet-4-5 & 93.22 & 78.67 & 64.00 & 61.69 & 45.17 & 18.67 \\
Deepseek-V3.2 & 90.00 & 60.00 & 65.50 & 29.02 & 20.50 & 13.00 \\
GLM-4.7 & 85.67 & 66.00 & 70.50 & 19.45 & 29.00 & 12.00 \\
\midrule
Ours & 85.00 & 44.00 & 58.50 & 5.50 & 24.50 & 8.00 \\
\bottomrule
\end{tabular}
\caption{The ability of CTI analysis benchmarked by AthenaBench-Mini. We compute all the results with greedy decoding.}
\label{tab:cti}
\end{table}

\paragraph{AthenaBench-Mini} AthenaBench~\citep{alam2025athenabench} is an updated benchmark of CTIBench~\citep{alam2024ctibench} for CTI analysis evaluation. The benchmark includes 6 distinct tasks of CTI reasoning: (1) CTI Knowledge Test (CTK), (2) Root Cause Mapping (RCM), (3) Vulnerability Severity Prediction (VSP), (4) Threat Actor Attribution (TAA), (5) Risk Mitigation Strategy (RMS), and (6) Attack Technique Extraction (ATE). We use the public AthenaBench-Mini set and report the results in \autoref{tab:cti}. The results suggest that \model achieve comparable performances to other open frontier models like Deepseek-V3.2 and GLM-4.7, but still fall short on the tasks related to root cause mapping and threat actor attribution. We plan to improve these tasks by adding more pre-training data on CTI analysis.

\begin{table}[ht]
\centering
\begin{tabular}{lcccc|cccc}
\toprule
Model & Acc & Precision & Recall & F1 & P-C $\downarrow$ & P-V $\downarrow$ & P-B $\downarrow$ & P-R $\downarrow$ \\
\midrule
Claude-Opus-4-5 & 52.73 & 52.42 & 59.02 & 55.53 & 9.02 & 50.06 & 37.28 & 3.63 \\
Claude-sonnet-4-5 & 52.57 & 51.44 & 91.80 & 65.93 & 8.52 & 83.38 & 4.93 & 3.17 \\
Deepseek-V3.2 & 50.09 & 51.27 & 4.89 & 8.91 & 3.51 & 1.00 & 91.60 & 3.88 \\
GLM-4.7 & 53.22 & 53.24 & 58.62 & 55.80 & 20.55 & 38.60 & 26.57 & 12.53 \\
\midrule
Ours & 48.33 & 48.54 & 54.54 & 51.37 & 0.88 & 53.01 & 41.29 & 4.64 \\
\bottomrule
\end{tabular}
\caption{The function-level vulnerability detection capability evaluated by PrimeVul-Paired. We note that Pair-wise Correct Prediction (P-C), Pair-wise Vulnerable Prediction (P-V), Pair-wise Benign Prediction (P-B), and Pair-wise Reversed Prediction (P-R) are used to evaluate the model's predictions on textually similar code pairs as single entities~\cite{}. We compute all the results with greedy decoding.}

\label{tab:primevul_paired_scores}
\end{table}

\paragraph{PrimeVul-Paired} To evaluate how well the models perform on vulnerability detection, we use the Paired set of PrimeVul~\citep{ding2024vulnerability}. Unlike the full set having significantly imbalanced vulnerable and benign samples, the Paired set has balanced labels and consists of textually similar code pairs where one function is vulnerable and the other is benign. This design allows for more rigorous evaluation of the model's ability to distinguish subtle differences that determine vulnerability status. As shown in \autoref{tab:primevul_paired_scores}, \model achieves the lowest P-C score, indicating superior performance in correctly identifying both vulnerable and benign functions within similar code contexts. While our model shows competitive F1 score and overall performance, it demonstrates significantly better paired prediction consistency compared to all baselines.

\begin{table}[!h]
    \centering
    \resizebox{\linewidth}{!}{
    \begin{tabular}{l cccc cc}
    \toprule
    \multirow{2}{*}{Model} & \multicolumn{4}{c}{SecCodeBench} & \multicolumn{2}{c}{CWEval} \\ \cmidrule(lr){2-5} \cmidrule(lr){6-7}
          & Gen w/o Hint & Gen w/ Hint & Fix w/o Hint & Fix w/ Hint & func@1 & func-sec@1 \\ 
    \midrule
    Claude-Opus-4-5 & 52.5 & 73.2 & 75.2 & 83.9  & 92.27 & 74.75 \\
    Claude-sonnet-4-5 & 43.4 & 57.6 & 62.2 & 76.4  & 91.55 & 71.72 \\
    Deepseek-V3.2 & 43.1 & 50.2 & 50.0 & 65.8  & 83.53 & 54.71 \\
    GLM-4.7 & 29.4 & 56.0 & 49.8 & 64.9  & 72.44 & 46.39 \\
    \midrule
    Ours & 61.2 & 69.5  & 76.4 & 83.7  & 80.17 & 56.32 \\
    \bottomrule
    \end{tabular}}
    \caption{The secure coding capability measured by SecCodeBench and CWEval. We note that SecCodeBench scores are derived from pass@$k$ computation with security severity weighting, with (and without) the hints that ask models to behave securely. In addition, func@1 and func-sec@1 are computed based on the definition of pass@$k$, with the random sampling of $n = 10$ and temperature of $0.8$. }
    \label{tab:secure_coding}
\end{table}

\paragraph{SecCodeBench and CWEval} We consider secure coding important in daily practice, where models help avoid and mitigate software vulnerabilities. Typically, there are two main scenarios: code generation and vulnerability repair. We focus on two existing benchmarks: SecCodeBench\footnote{\url{https://github.com/alibaba/sec-code-bench}} and CWEval~\citep{peng2025cweval}. SecCodeBench consists of 53 Java coding tasks, the majority of which are derived from anonymized, real-world historical vulnerabilities at Alibaba, covering generation and repair scenarios both with and without security hints. CWEval is a multilingual secure code generation benchmark, with a focus on both functionality and security. 
From \autoref{tab:secure_coding}, we observe that \model achieves competitive security performance on SecCodeBench across both generation and repair tasks. Notably, it maintains high scores even without security hints, particularly outperforming Claude-Opus-4-5 on generation (61.2 vs. 52.5), indicating robust inherent security awareness in default code generation.
For CWEval, we report func@1 and func-sec@1 for functionality and security, respectively. While Claude models achieve higher func@1 scores, \model demonstrates competitive secure code generation capability with func-sec@1 of 56.32\%, outperforming Deepseek-V3.2 and GLM-4.7, indicating a reasonable balance between generating functional and secure code.

\end{document}